# Training Affective Computer Vision Models by Crowdsourcing Soft-Target Labels

Peter Washington*, Department of Bioengineering, Stanford University, peterwashington@stanforrd.edu

Haik Kalantarian, Department of Pediatrics (Systems Medicine), Stanford University, haik.kalantarian@gmail.com

Jack Kent, Department of Pediatrics (Systems Medicine), Stanford University, jackkent@stanford.edu

Arman Husic, Department of Pediatrics (Systems Medicine), Stanford University, ahusic@stanford.edu

Aaron Kline, Department of Pediatrics (Systems Medicine), Stanford University, akline@stanford.edu

Emilie Leblanc, Department of Pediatrics (Systems Medicine), Stanford University, emilie.leblanc@stanford.edu

Cathy Hou, Department of Computer Science, Stanford University, cathyhou@stanford.edu

Cezmi Mutlu, Department of Electrical Engineering, Stanford University, cezmi@stanford.edu

Kaitlyn Dunlap, Department of Pediatrics (Systems Medicine), kaiti.dunlap@stanford.edu

Yordan Penev, Department of Pediatrics (Systems Medicine), Stanford University, ypenev@stanford.edu

Nate Stockham, Department of Neuroscience, Stanford University, stockham@stanford.edu

Brianna Chrisman, Department of Bioengineering, Stanford University, briannac@stanford.edu

Kelley Paskov, Department of Biomedical Data Science, Stanford University, kpaskov@stanford.edu

Jae-Yoon Jung, Department of Pediatrics (Systems Medicine), Stanford University, jaeyjung@stanford.edu

Catalin Voss, Department of Computer Science, Stanford University, catalin@cs.stanford.edu

Nick Haber, Graduate School of Education, Stanford University, nhaber@stanford.edu

Dennis P. Wall[*], Departments of Pediatrics (Systems Medicine), Biomedical Data Science, and Psychiatry and Behavioral Sciences, Stanford University, dpwall@stanford.edu

*Corresponding authors: Peter Washington (peterwashington@stanford.edu), Dennis P. Wall (dpwall@stanford.edu)

**ABSTRACT**

Emotion classifiers traditionally predict discrete emotions. However, emotion expressions are often subjective, thus requiring a method to handle subjective labels. We explore the use of crowdsourcing to acquire reliable soft-target labels and evaluate an emotion detection classifier trained with these labels. We center our study on the Child Affective Facial Expression (CAFE) dataset, a gold standard collection of images depicting pediatric facial expressions along with 100 human labels per image. To test the feasibility of crowdsourcing to generate these labels, we used Microworkers to acquire labels for 207 CAFE images. We evaluate both unfiltered workers as well as workers selected through a short crowd filtration process. We then train two versions of a classifiers on soft-target CAFE labels using the original 100 annotations provided with the dataset: (1) a classifier trained with traditional one-hot encoded labels, and (2) a classifier trained with vector labels representing the distribution of CAFE annotator responses. We compare the resulting softmax output distributions of the two classifiers with a 2-sample independent t-test of L1 distances between the classifier's output probability distribution and the distribution of human labels. While agreement with CAFE is weak for unfiltered crowd workers, the filtered crowd agree with the CAFE labels 100% of the time for many emotions. While the F1-score for a one-hot encoded classifier is much higher (94.33% vs. 78.68%) with respect to the ground truth CAFE labels, the output probability vector of the crowd-trained classifier more closely resembles the distribution of human labels (t=3.2827, p=0.0014). Reporting an emotion probability distribution that accounts for the subjectivity of human interpretation. Crowdsourcing, including a sufficient filtering mechanism, is a feasible solution for acquiring soft-target labels.

**INTRODUCTION**

Machine learning models which predict human emotion from images of facial expressions are increasingly used in interactive systems [5, 33-34, 51, 83] and applications such as multimodal sentiment analysis [42, 67-68], healthcare [74, 100], and autonomous vehicles [23]. Emotion recognition is traditionally modeled as a classification problem, where the model predicts a discrete emotion category. However, facial expressions are often ambiguous [15, 104 107], and it is often not ideal for a machine learning model to output a single class for a subjective label. Fortunately, most supervised learning methods output a probability distribution over all possible classes. Sometimes, the affective computing system will visualize this distribution to the user [33]. Examples include commercial emotion detection services like Affectiva [55-56] and autonomous vehicles displays [83]. In a large number of use cases, however, only the class with the highest probability is visualized [34, 51].

While the paradigm of training a model with a discrete one-hot encoded label and predicting a probability distribution is reasonable when the training data have indisputable labels, images of facial expressions can have ambiguous labels or even multiple correct labels simultaneously, and the label should ideally represent this inherent uncertainty. Soft-target labeling, where the training labels represent a probabilistic distribution rather than a one-hot encoded label, is an established solution to this issue. Training with soft-target labels results in classifiers which predict probability distributions representative of the soft-target labels [3, 25, 101]. We hypothesize that crowdsourcing can generate distributions which mirror those generated in a lab setting.

Here, we explore the use of crowdsourcing to acquire a distribution of labels for images with ambiguous or multiple classes (we call these "subjective labels"). We first describe the acquisition of crowdsourced labels for four representative images which we display to the reader along with the distribution of crowd responses to demonstrate the phenomena of subjective labels in affective computing. We then crowdsource the labeling of a subset of the Child Affective Facial Expression (CAFE) dataset, a collection of emotive images of children which conveniently comes with 100 independent human annotations per image. We next show that the crowdsourced distribution mirrors the original CAFE distribution, validating the feasibility of crowdsourcing for generating a reliable and representative distribution of human labels for an image. Finally, we compare the performance of two versions of a convolutional neural network (CNN) trained on CAFE: one with traditional one-hot encoded vectors and the other with soft-target labels based on CAFE annotator responses. We find that the classifier trained with soft targets results in classifier predictions that much more closely mirror the true human distribution on independent subjects not included in the training set. We hope that this work will be of use to designers and developers of machine learning models for affective computing systems who wish to provide probabilistic outputs to the end user.

## RELATED WORK

While crowdsourcing and soft-target labels have been studied in affective computing, we are the first to explore the feasibility of using crowdsourcing to acquire reliable soft-target labels for computer vision emotion detection. We describe related work below.

### Facial emotion detection

Facial emotion detection is a key challenge for machine learning. For intelligent machines to convincingly pass the Turing test [80], an understanding of human emotion is crucial. There has been a strong body of machine learning literature for detecting human affect from a variety of data streams, including audio [67-68, 103], text [5, 48], images [75], and video [28, 84-85]. Here, we focus on image-based emotion detection from facial expressions.

Fundamental to a successful computer vision approach for affective computing is the feature representation of the image, and there are several approaches to engineering such features. A common approach is to extract facial keypoints and use a feature representation consisting of the coordinates of the keypoints [16, 28, 60, 71]. This approach works well when the dataset is small, as the representation itself is compact and therefore amenable to lightweight learning approaches such as logistic regression, support vector machines, and decision trees. Another feature extraction approach, CNNs, can automatically learn relevant nonlinear feature maps. CNNs oftentimes result in superior performance to other methods when the dataset is sufficiently large [24, 89].

### Emotion detection with subjective labels

Paul Ekman posited that there are seven fundamental human emotions which are universal across cultures and geographic boundaries: happy, sad, surprise, anger, fear, disgust, and contempt [21-22]. However, these expressions are not mutually exclusive. Du et al. discussed the existence of *compound emotions*, or combinations of existing emotions to form new ones [17]. Examples of compound emotions include "happily surprised", "fearfully surprised", and "fearfully disgusted". Through smartphone sensing, Zhang et al. found that pairs of emotions which are often presented simultaneously include (happy, surprised), (sad, disgust), and (sad, fear) [105]. This issue has been explored for emotional speech [15, 107]. While some emotions may be jointly expressed, others may be singular yet ambiguous. The issue of subjectivity in training labels, whether due to ambiguous labels or multiple correct labels, has been documented in the fields of digital health and affective computing in particular [54, 57, 69, 82, 104].

The topic of subjective labels in affective computing datasets containing speech and audio data has been explored in prior work. Mower et al. represent emotion labels at the granularity of utterances, thereby representing a time profile of how the dominant emotion in speech changes quickly over time [58-59]. Fujioka alternates between updating neural network parameters and updating sample importance parameters at each training iteration [26]. Ando et al. utilize soft-target training, where the emotion labels are based on the proportion of human annotations instead of the traditional one-hot encoding [3].

Soft-target training is a general machine learning method for handling subjective training labels. This approach is particularly desirable when multiple labels are acquired per image. Classification of soft labels can be beneficial because they can account for inherit subjectivity in labels and are robust against random noise [79]. A variation of this method is a soft loss function, which consists of subtracting the minimum between-class distance from the maximum within-class distance [99]. Soft-target and loss training have been shown to outperform hard target training (one-hot encoding) when the training goal is to produce an output distribution like the distribution of annotator labels [65], and this phenomenon has been observed across several datasets and tasks [81].

The issue of subjective labels has also been explored in multimodal sentiment analysis, where the goal is to predict sentiment from multiple data streams [42, 67-68], including affect-enriched videos. Chaturvedi et al. created a fuzzy classifier for predicting the degree to which several emotions are expressed in a particular image [6]. Another approach is to predict the amount of valence and arousal displayed on continuous axes (regression) rather than predicting categories (classification) [62, 64, 73, 102, 106].

**Crowdsourcing with subjective labels**

There are several bodies of work which describe approaches to handling crowdsourced labels. Kairam and Heer hypothesize that there are intrinsic but valid differences between crowd workers when labeling data points and therefore categorize workers by their labeling patterns [40]. Other examples of categorizing workers include measuring sample informativeness, active and cooperative learning strategies, and controlling for labeler trustworthiness metrics [69].

There have been other statistical learning techniques beyond the soft-target labeling discussed above which have been successful with crowdsourced labels. Rodrigues and Pereira add an extra "crowd layer" at the end of a traditional CNN architecture trained to predict the outputs of each labeler individually and therefore the biases of crowd workers [70].

Crowdsourcing has been used to acquire emotion labels of images. Korovina et al. found that crowd workers labeling discrete emotion categories on a color wheel had low agreement scores (Kappa value less than 0.15) [46] while consistency between workers when labeling valence and arousal was much stronger [45].

**METHODS**

**Acquiring Crowd Labels for CAFE Images**

We use the CAFE dataset [52-53], which is the largest public dataset of front-facing images depicting children emoting. CAFE is used as a benchmark in several affective computing publications [52-53] and is the standard evaluation dataset for pediatric affective computing. CAFE was originally labeled by 100 untrained human raters, and the raw distribution of 100 human labels per image are provided along with the ground truth labels. For example, the first image in the dataset was labeled as "angry" by 62% of raters and as "disgusted" by 25% of raters. All other emotions received 5% or fewer labels. One could hypothesize from these numbers that the image looks "mostly angry" with "some disgust". Manually inspecting the image reveals a facial expression which could reasonably be categorized as either "angry" or "disgusted" depending on the context. (CAFE images are protected by copyright and cannot be republished, so we refer to this image by its filename in the publicly available dataset: *F-AA-01_052-Angry.jpg*).

To validate the capability of crowdsourcing to produce a reliable ground truth label distribution, we crowdsourced the task of labeling CAFE images and compared the resulting crowd-generated distribution to the distribution reported in the original CAFE dataset. All crowdsourcing was conducted on Microworkers.com, a crowdsourcing platform similar to Amazon Mechanical Turk [63] but with a more globally representative pool of workers [32]. Each task consisted of labeling one of seven emotion categories (happy, sad, surprised, angry, fearful, disgusted, and neutral) for a subset of images in CAFE. We chose to limit rater labels to absolute ratings (one-hot representations) because we wanted to capture the relative weighting of each emotion within an image. Because humans are notoriously poor at precisely quantifying relative contributions of individual components in mixed representations, especially in the case of human emotion recognition [8-9, 27], we asked each rater to only provide the most salient emotion according to their interpretation. By acquiring labels from 100 independent crowd workers per image, each providing their vote for the most prominent emotion, we created a representation describing the between-subject subjectivity of the emotion expressed in the image.

We acquired labels for 131 randomly selected images from CAFE and we solicited 100 crowd labels per image. We manually checked each label for correctness, and workers with consistently high-quality labels were recruited for additional labeling tasks for 76 separate images. Here, "high-quality" means that the authors could potentially agree with the label (e.g., a "happy" label for a clearly "sad" image would not be accepted, but a "fearful" label for a "fearfully surprised" expression would be accepted). Our goal when excluding workers without consistently "high-quality" labels was to filter out crowd workers who were answering randomly to receive payment, as this is a common issue in crowdsourcing [2, 4, 10, 49]. We analyzed both the filtered and unfiltered worker labels on different sets of CAFE images to measure the possibility that filtering workers could mask ambiguity of the labels.

All crowdsourcing tasks were approved by the Institutional Review Board (IRB) of Stanford University. All workers were required to sign an electronic consent form approved by the IRB before participating in the task.

**Training and Testing with Crowd Probability Distributions**

Traditionally, multi-class models are trained with categorical cross-entropy loss, where $\sum$ is the summation operator, $C$ is the number of classes, $p_i$ is the ground truth probability of class $i$, and $q_i$ is the classifier prediction for class $i$:

$$-\sum_{i}^{C} p_i \log(q_i)$$

When the true classes are indisputable, which is the usual assumption for classification, then the ground truth probability distribution $p_i$ is a one-hot encoding (i.e., a probability of 1 for the "true" class and a probability of 0 for all other classes). In the case of subjective classes where the true label may consist of a weighted combination of multiple classes, like in emotion datasets where complex emotions are present, we hypothesize that providing soft-target labels instead of one-hot encodings will result in classifier predictions for separate human subjects which resemble the human annotator response distribution.

We trained a machine learning model using two sets of image labels: (1) the original CAFE labels as one-hot encoded vectors and (2) soft-target vectors representing the distribution of 100 human responses from the original CAFE dataset. We held out all images from 5 randomly selected child subjects from CAFE (F-AA-01, F-EA-39, M-LA-08, M-AA-11, and F-LA-13, corresponding to one female African American, one female European American, one male Latin American, one male African American, and one female Latin American) and used these as test set images. The rest of the images were used to train the classifier. 1,141 images (196 angry, 180 disgusted, 135 fearful, 206 happy, 222 neutral, 103 sad, and 99 surprised) were used in the train set and 51 images (9 angry, 11 disgusted, 5 fearful, 9 happy, 8 neutral, 5 sad, and 4 surprised) were used in the test set.

We transfer learned on a ResNet-152 [31] CNN pretrained on ImageNet [13]. We trained each neural network using the Keras framework [7] with a TensorFlow [1] backend for 100 epochs with a batch size of 16 and a learning rate of 0.0003 using Adam optimization [43]. To increase generalization of the training process and reduce overfitting, we applied the following data augmentation strategies: a rotation range of 7 degrees, a zoom range of 15%, a shear range of 5%, a brightness range of 70% to 130%, and horizontal flipping.

## RESULTS

### Demonstration of Subjective Emotions

The methods described here are not specific to CAFE. We focus on CAFE in this paper as a case study of a popular affective computing dataset and as a dataset which provides ground truth labels for many human annotators (100) per image. However, CAFE images are subject to copyright and cannot be republished. To provide the reader with visual examples of facial expressions with large numbers of crowd annotations per image, we display free-to-republish images in Figure 1. For each image, we acquired 200 crowdsourced labels from Microworkers.com, as described above.

Figures 1A shows an image that could be labeled as either angry or disgusted, and Figure 1C shows an image that is possibly angry, fearful, surprised, or some combination of the 3. Further context is required to reach full confidence about the true classes. Figure 1B displays a compound emotion, where the individual appears to be "fearfully surprised". Assigning only a single category to the image would be misleading. Figure 1D depicts a situation where it is unclear whether the individual's neutral face looks sad or if that individual is making a sad face. In cases like this, a personalized emotion recognition model would likely be required.

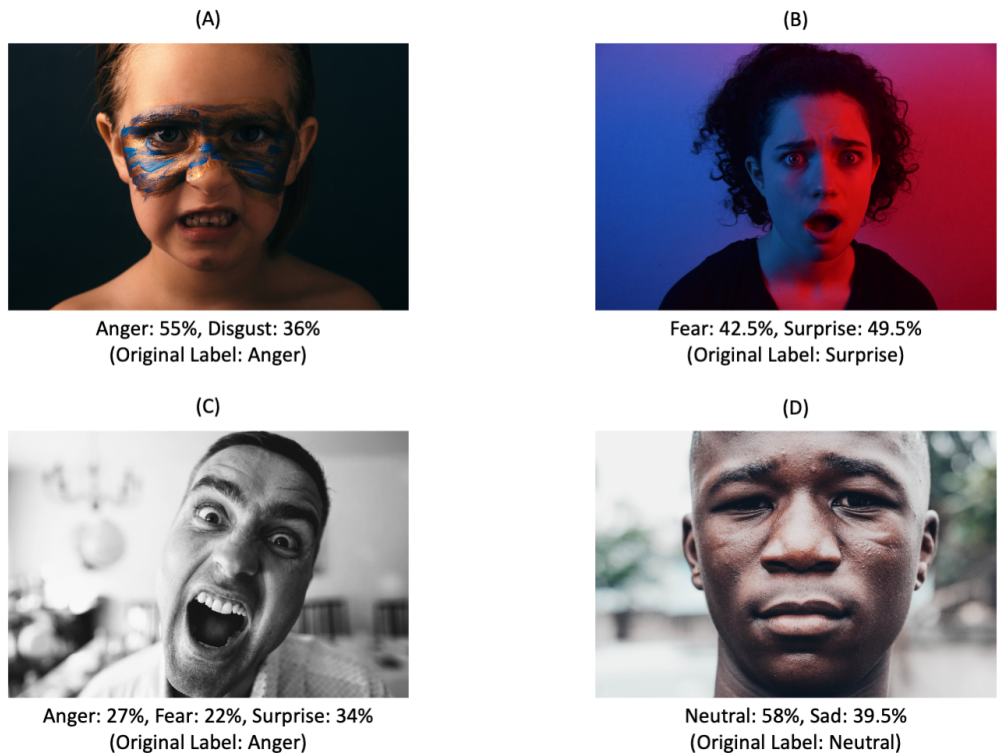

Figure 1: Examples of images with subjective emotion labels. We acquired 200 crowdsourced annotations for each image above. Percentages of labels for each emotion category are displayed under the image for all emotions receiving at least 10% of votes. (A) This expression could be anger or disgust. (B) This expression is a compound expression of fear and surprise. (C) Depending on context, this image could be anger, fear, or surprise (or some combination). (D) It is unclear if this face is neutral or sad, highlighting a need for personalized emotion recognition techniques.

We also quantified the subjectivity of images in CAFE. We measured the number of images with 80% of annotations represented with the top-N most frequent labels for N ranging from 1 to 5 inclusive (histogram in Figure 2). We see that while many emotions do not contain much subjectivity (N=1), most images are either ambiguous between or compound with 2 or more emotions. When the cutoff is increased to 90% (Figure 3), the number of subjective labels increases further.

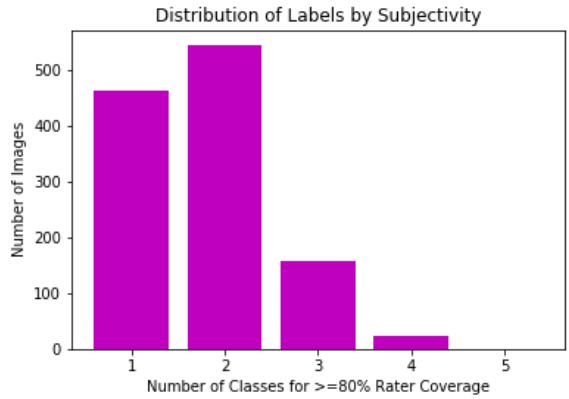

Figure 2: Distribution of labels by subjectivity. Histogram of the number of highest-voted classes required to reach greater than or equal to 80% rater coverage for each image.

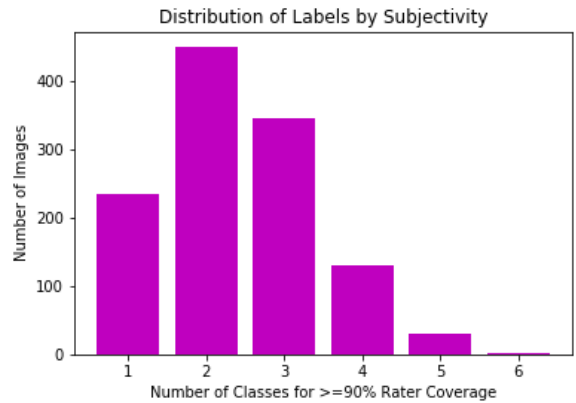

Figure 3: Distribution of labels by subjectivity. Histogram of the number of highest-voted classes required to reach greater than or equal to 90% rater coverage for each image.

**Comparison of CAFE Labels and Crowd Performance**

When looking at the majority consensus label, the filtered crowd agreed with the CAFE labels 100% of the time for happy, neutral, sad, and surprise. There was 90% agreement for disgust, 75% agreement for anger, and 50% agreement for fear. When combining commonly confused labels into one class ("anger + disgust" and "fear + surprise"), the filtered crowd agreed with the CAFE labels 100% of the time for happy, neutral, sad, and "fear + surprise" and 88.8% for "anger + disgust".

Table 1: CAFE original annotator distribution vs. *filtered* worker's distribution for subject F-AA-15 in CAFE.

| Image | CAFE Labeler Distribution (count) | Crowdsourced Labeler Distribution (count) |
|---|---|---|
| | Anger, Disgust, Fear, Happy, Neutral, Sad, Surprised | |
| 9990-angry_F-AA-15.jpg | 30, 37, 15, 8, 0, 8, 2 | 7, 3, 0, 4, 0, 0, 0 |
| 10108-angryopen_F-AA-15.jpg | 29, 6, 35, 1, 1, 23, 5 | 2, 2, 4, 0, 0, 6, 0 |
| 10194-disgust_F-AA-15.jpg | 3, 86, 3, 2, 1, 5, 0 | 2, 10, 0, 0, 0, 2, 0 |
| 10288-disgustwithtongue_F-AA-15.jpg | 3, 91, 0, 3, 2, 0, 1 | 1, 6, 1, 5, 0, 0, 1 |
| 10383-fearful_F-AA-15.jpg | 2, 1, 82, 2, 1, 6, 6 | 0, 1, 10, 0, 0, 0, 3 |
| 10461-fearfulopen_F-AA-15.jpg | 2, 3, 58, 2, 3, 1, 31 | 0, 0, 5, 0, 0, 0, 9 |
| 10526-happy_F-AA-15.jpg | 1, 0, 0, 96, 2, 1, 0 | 0, 0, 0, 14, 0, 0, 0 |
| 10739-neutral_F-AA-15.jpg | 1, 0, 1, 1, 89, 7, 1 | 0, 0, 0, 0, 14, 0, 0 |
| 10867-neutralopen_F-AA-15.jpg | 2, 2, 10, 1, 33, 0, 52 | 0, 0, 0, 0, 7, 0, 7 |
| 10967-sad_F-AA-15.jpg | 3, 3, 6, 1, 2, 85, 0 | 2, 0, 0, 0, 0, 12, 0 |
| 11027-sadopen_F-AA-15.jpg | 0, 5, 22, 0, 0, 72, 1 | 0, 0, 3, 0, 0, 11, 0 |
| 11079-surprise_F-AA-15.jpg | 1, 0, 23, 0, 2, 0, 74 | 0, 0, 1, 0, 0, 0, 13 |

By contrast, the unfiltered crowd workers did not agree as strongly with the CAFE labels when looking at the majority consensus, highlighting the need for quality control measures when crowdsourcing emotion annotations. There was 100% agreement for surprised, 93.3% agreement for happy, 83.3% agreement for sad, 76.9% agreement for disgusted, 64.3% agreement for angry, 61.5% agreement for neutral, and 30.8% agreement for fearful.

Table 2: CAFE original annotator distribution vs. *unfiltered* crowdsourced class distribution for subject F-AA-93 in CAFE.

| Image | CAFE Labeler Distribution (count) | Crowdsourced Labeler Distribution (count) |
|---|---|---|
| | Anger, Disgust, Fear, Happy, Neutral, Sad, Surprised | |
| 9979-angry_F-AA-03.jpg | 89, 4, 0, 0, 0, 4, 3 | 78, 31, 2, 8, 3, 2, 3 |
| 10100-angryopen_F-AA-03.jpg | 16, 0, 36, 5, 1, 2, 40 | 17, 3, 38, 15, 0, 0, 54 |
| 10184-disgust_F-AA-03.jpg | 17, 41, 2, 12, 19, 8, 1 | 12, 75, 1, 10, 25, 3, 1 |
| 10280-disgustwithtongue_F-AA-03.jpg | 19, 77, 0, 0, 2, 1, 1 | 19, 85, 2, 13, 6, 0, 2 |
| 10375-fearful_F-AA-03.jpg | 2, 4, 49, 13, 2, 3, 27 | 10, 12, 27, 15, 16, 3, 44 |
| 10454-fearfulopen_F-AA-03.jpg | 0, 1, 27, 4, 1, 0, 67 | 1, 0, 43, 3, 0, 0, 80 |
| 10515-happy_F-AA-03.jpg | 1, 0, 0, 98, 1, 0, 0 | 0, 4, 0, 113, 8, 1, 1 |
| 10635-happyopen_F-AA-03.jpg | 1, 2, 0, 94, 1, 0, 2 | 0, 0, 0, 126, 0, 0, 1 |
| 10730-neutral_F-AA-03.jpg | 3, 2, 0, 2, 73, 19, 1 | 2, 1, 0, 0, 77, 47, 0 |
| 10858-neutralopen_F-AA-03.jpg | 4, 4, 4, 2, 17, 3, 66 | 1, 7, 21, 1, 10, 5, 82 |
| 10960-sad_F-AA-03.jpg | 2, 1, 2, 1, 2, 92, 0 | 8, 9, 3, 0, 4, 103, 0 |
| 11021-sadopen_F-AA-03.jpg | 1, 6, 21, 15, 13, 30, 14 | 9, 21, 24, 14, 12, 43, 4 |
| 11068-surprise_F-AA-03.jpg | 2, 1, 13, 20, 1, 0, 63 | 1, 1, 12, 30, 0, 1, 82 |

Tables 1 and 2 compare the distribution of labels from the original CAFE labelers as well as the filtered and unfiltered crowd workers (respectively) for a single image. In both the filtered and unfiltered cases, the distributions qualitatively mirror each other in terms of their peaks. In all cases, peaks which appear in the CAFE annotator distribution also appear in the crowd distribution, and vice versa. However, these distributions are noisy, and the relationship between the peaks cannot be guaranteed (e.g., if "anger" has more labels than "disgust" for CAFE annotators, "disgust" may have more ratings for crowd annotators). The generated crowd distributions must therefore be regarded as a noisy approximation to the true probability distribution, and further work should account for this noise in the label representation.

**Training and Testing with Crowd Probability Distributions**

We evaluate the models with F1-score rather than accuracy because CAFE is not a balanced dataset. When training with the one-hot encoded labels, the F1-score on the held-out test set is 94.33%. We emphasize that this high performance is misleading due to the ambiguity of the ground truth labels. When training with vectors representing the distribution of human labels, the F1-score on the held-out test-set is 78.68%. While the F1 score is lower when training with human distribution labels, the distribution of emotion predictions much more closely resembles the distribution of human labels for the distribution-trained classifier. For many applications of affective computing, having a representative label distribution is more important than absolute accuracy. The mean L1 distance between the human label distribution for the test set and distribution-trained classifier is 0.3727 (SD=0.3000); the mean L1 distance between the human label distribution and one-hot encoding-trained classifier is 0.6078 (SD=0.4143). The difference in L1 distances between these two groups is statistically significant according to an independent 2-sample t-test ($t=3.2827$, $p=0.0014$). To visualize this difference, Figure 4 compares the true emotion distribution with the emitted distribution of each of the two classifiers for 3 representative images in the test set with subjective labels.

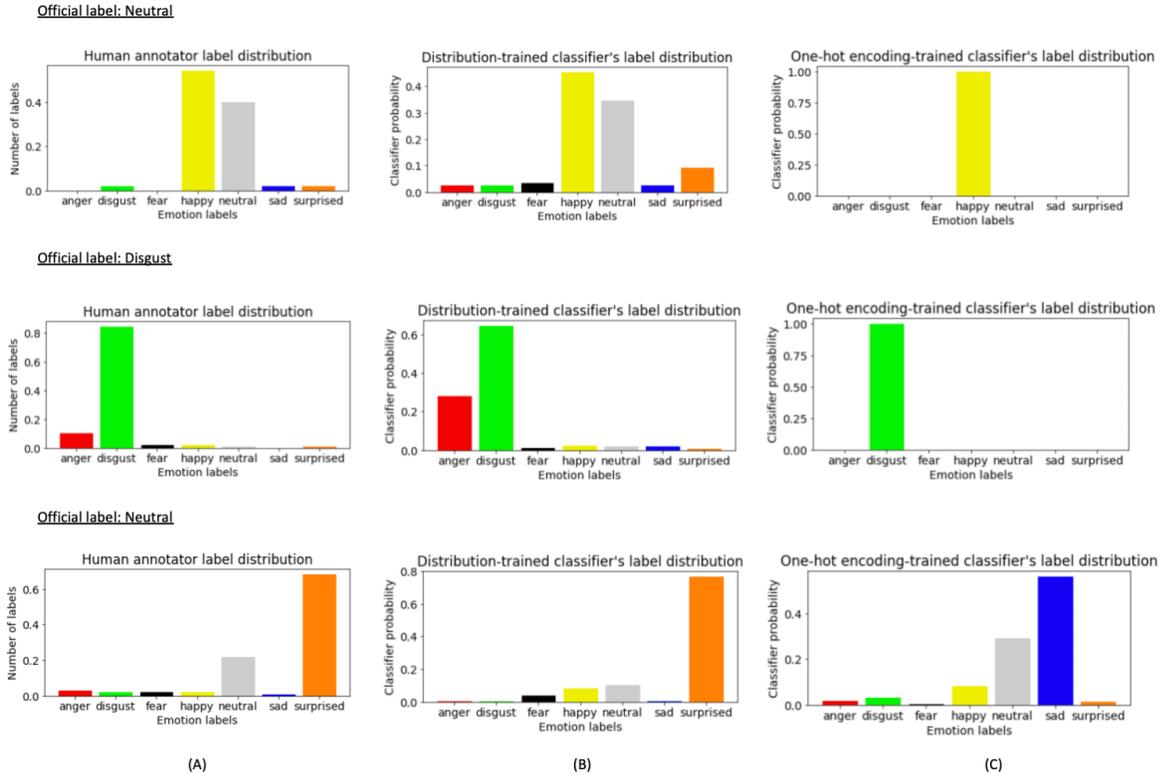

Figure 4: Comparison of human label distribution (A) and the predicted probability of a classifier trained with a probability vector representing the variety of human labels (B) vs. a one-hot encoded vector for images with subjective emotion labels (C).

**DISCUSSION**

Interaction designers and developers of affective computing systems should consider whether soft or hard targets is the most appropriate label representation for training an affective computer vision classifier for a particular application and dataset. Affective computer vision models which are optimized for understanding the potentially diverse range of human interpretations of emotion can be used in several applications of interactive systems, such as AI-powered systems which aid individuals with autism and other developmental delays [11-12, 14, 29, 35-39, 44, 66, 72, 84-85, 95-96], e-learning systems [41, 78], or at-home diagnostic screening tools for psychiatry conditions [18-20, 30, 47, 50, 76-77, 86-88, 90-94, 97].

There are several limitations of this work. This study was performed on a single dataset. For these results to generalize to other types of images, including for crowdsourced soft-target label generation in domains outside of emotion recognition, other datasets must be explored. Another limitation is that we did not record or account for potential biases in the quality control steps for filtering the crowd. Further study into how differing crowd quality mechanisms affect the result would be interesting, as the data label quality can drastically affect of a machine learning algorithm. A final limitation is that we did not have a reliable method to disentangle compound emotions from ambiguous labels.

By acquiring labels from 100 independent crowd workers per image, each providing their vote for the most prominent emotion, we created a representation describing the between-subject subjectivity of the emotion expressed in the image. This representation notably obfuscates within-subject subjectivity, and an alternative which should be studied in future work is to ask each rater to provide multiple selections through a semantical scale, as in Korovina et al. [45-46].

## CONCLUSION

For many applications of affective computing, reporting an emotion probability distribution that accounts for the subjectivity of human interpretation can be more important than traditional machine learning metrics. Crowdsourcing is a feasible solution for acquiring soft-target labels provided a sufficient filtering mechanism for selecting reliable crowd workers.

## ACKNOWLEDGMENTS

This work was supported in part by funds to DPW from the National Institutes of Health (1R01EB025025-01, 1R21HD091500-01, 1R01LM013083, 1R01LM013364), the National Science Foundation (Award 2014232), The Hartwell Foundation, Bill and Melinda Gates Foundation, Coulter Foundation, Lucile Packard Foundation, the Weston Havens Foundation, and program grants from Stanford's Human Centered Artificial Intelligence Program, Stanford's Precision Health and Integrated Diagnostics Center (PHIND), Stanford's Beckman Center, Stanford's Bio-X Center, Predictives and Diagnostics Accelerator (SPADA) Spectrum, Stanford's Spark Program in Translational Research, Stanford mediaX, and Stanford's Wu Tsai Neurosciences Institute's Neuroscience: Translate Program. We also acknowledge generous support from David Orr, Imma Calvo, Bobby Dekesyer and Peter Sullivan. P.W. would like to acknowledge support from Mr. Schroeder and the Stanford Interdisciplinary Graduate Fellowship (SIGF) as the Schroeder Family Goldman Sachs Graduate Fellow.

## COMPLIANCE WITH ETHICAL STANDARDS

Funding: This study was funded by the National Institutes of Health (1R01EB025025-01, 1R21HD091500-01, 1R01LM013083, 1R01LM013364), the National Science Foundation (Award 2014232), The Hartwell Foundation, Bill and Melinda Gates Foundation, Coulter Foundation, Lucile Packard Foundation, the Weston Havens Foundation, and program grants from Stanford's Human Centered Artificial Intelligence Program, Stanford's Precision Health and Integrated Diagnostics Center (PHIND), Stanford's Beckman Center, Stanford's Bio-X Center, Predictives and Diagnostics Accelerator (SPADA) Spectrum, Stanford's Spark Program in Translational Research, Stanford mediaX, and Stanford's Wu Tsai Neurosciences Institute's Neuroscience: Translate Program. We also acknowledge generous support from David Orr, Imma Calvo, Bobby Dekesyer and Peter Sullivan. P.W. would like to acknowledge support from Mr. Schroeder and the Stanford Interdisciplinary Graduate Fellowship (SIGF) as the Schroeder Family Goldman Sachs Graduate Fellow.

Informed consent: Informed consent was obtained from all individual participants (crowd workers on Microworkers.com) included in the study.

**REFERENCES**

[1] Abadi, Martín, Paul Barham, Jianmin Chen, Zhifeng Chen, Andy Davis, Jeffrey Dean, Matthieu Devin et al. "Tensorflow: A system for large-scale machine learning." In *12th {USENIX} symposium on operating systems design and implementation ({OSDI} 16)*, pp. 265-283. 2016.

[2] Allahbakhsh, Mohammad, Boualem Benatallah, Aleksandar Ignjatovic, Hamid Reza Motahari-Nezhad, Elisa Bertino, and Schahram Dustdar. "Quality control in crowdsourcing systems: Issues and directions." *IEEE Internet Computing* 17, no. 2 (2013): 76-81.

[3] Ando, Atsushi, Satoshi Kobashikawa, Hosana Kamiyama, Ryo Masumura, Yusuke Ijima, and Yushi Aono. "Soft-target training with ambiguous emotional utterances for DNN-based speech emotion classification." In *2018 IEEE International Conference on Acoustics, Speech and Signal Processing (ICASSP)*, pp. 4964-4968. IEEE, 2018.

[4] Buchholz, Sabine, and Javier Latorre. "Crowdsourcing preference tests, and how to detect cheating." In *Twelfth Annual Conference of the International Speech Communication Association*. 2011.

[5] Cambria, Erik, Dipankar Das, Sivaji Bandyopadhyay, and Antonio Feraco. "Affective computing and sentiment analysis." In *A practical guide to sentiment analysis*, pp. 1-10. Springer, Cham, 2017.

[6] Chaturvedi, Iti, Ranjan Satapathy, Sandro Cavallari, and Erik Cambria. "Fuzzy commonsense reasoning for multimodal sentiment analysis." *Pattern Recognition Letters* 125 (2019): 264-270.

[7] Chollet, François. "Keras: The python deep learning library." *ascl* (2018): ascl-1806.

[8] Coolican, Jamesie, Gail A. Eskes, Patricia A. McMullen, and Erin Lecky. "Perceptual biases in processing facial identity and emotion." *Brain and Cognition* 66, no. 2 (2008): 176-187.

[9] Coren, Stanley, and James A. Russell. "The relative dominance of different facial expressions of emotion under conditions of perceptual ambiguity." *Cognition and Emotion* 6, no. 5 (1992): 339-356.

[10] Daniel, Florian, Pavel Kucherbaev, Cinzia Cappiello, Boualem Benatallah, and Mohammad Allahbakhsh. "Quality control in crowdsourcing: A survey of quality attributes, assessment techniques, and assurance actions." *ACM Computing Surveys (CSUR)* 51, no. 1 (2018): 1-40.

[11] Daniels, Jena, Jessey N. Schwartz, Catalin Voss, Nick Haber, Azar Fazel, Aaron Kline, Peter Washington, Carl Feinstein, Terry Winograd, and Dennis P. Wall. "Exploratory study examining the at-home feasibility of a wearable tool for social-affective learning in children with autism." *NPJ digital medicine* 1, no. 1 (2018): 1-10.

[12] Daniels, Jena, Nick Haber, Catalin Voss, Jessey Schwartz, Serena Tamura, Azar Fazel, Aaron Kline et al. "Feasibility testing of a wearable behavioral aid for social learning in children with autism." *Applied clinical informatics* 9, no. 1 (2018): 129.

[13] Deng, Jia, Wei Dong, Richard Socher, Li-Jia Li, Kai Li, and Li Fei-Fei. "Imagenet: A large-scale hierarchical image database." In *2009 IEEE conference on computer vision and pattern recognition*, pp. 248-255. Ieee, 2009.

[14] Deriso, David, Joshua Susskind, Lauren Krieger, and Marian Bartlett. "Emotion mirror: a novel intervention for autism based on real-time expression recognition." In *European Conference on Computer Vision*, pp. 671-674. Springer, Berlin, Heidelberg, 2012.

[15] Devillers, Laurence, Laurence Vidrascu, and Lori Lamel. "Challenges in real-life emotion annotation and machine learning based detection." *Neural Networks* 18, no. 4 (2005): 407-422.

[16] Dinculescu, Adrian, Andra Băltoiu, Carmen Strungaru, Livia Petrescu, Cristian Vizitiu, Alexandru Mandu, Nicoară Talpeș, and Vlad Văleanu. "Automatic Identification of Anthropological Face Landmarks for Emotion Detection." In *2019 9th International Conference on Recent Advances in Space Technologies (RAST)*, pp. 585-590. IEEE, 2019.

[17] Du, Shichuan, Yong Tao, and Aleix M. Martinez. "Compound facial expressions of emotion." *Proceedings of the National Academy of Sciences* 111, no. 15 (2014): E1454-E1462.

[18] Duda, Marlena, Jena Daniels, and Dennis P. Wall. "Clinical evaluation of a novel and mobile autism risk assessment." *Journal of autism and developmental disorders* 46, no. 6 (2016): 1953-1961.

[19] Duda, M., N. Haber, J. Daniels, and D. P. Wall. "Crowdsourced validation of a machine-learning classification system for autism and ADHD." *Translational psychiatry* 7, no. 5 (2017): e1133-e1133.

[20] Duda, M., R. Ma, N. Haber, and D. P. Wall. "Use of machine learning for behavioral distinction of autism and ADHD." *Translational psychiatry* 6, no. 2 (2016): e732-e732.

[21] Ekman, Paul. "Are there basic emotions?." (1992): 550.

[22] Ekman, Paul. "Basic emotions." *Handbook of cognition and emotion* 98, no. 45-60 (1999): 16.

[23] Eyben, Florian, Martin Wöllmer, Tony Poitschke, Björn Schuller, Christoph Blaschke, Berthold Färber, and Nhu Nguyen-Thien. "Emotion on the road—necessity, acceptance, and feasibility of affective computing in the car." *Advances in human-computer interaction* 2010 (2010).

[24] Fan, Yingruo, Jacqueline CK Lam, and Victor OK Li. "Multi-region ensemble convolutional neural network for facial expression

recognition." In *International Conference on Artificial Neural Networks*, pp. 84-94. Springer, Cham, 2018.

[25] Fang, Xi, Jiancheng Yang, and Bingbing Ni. "Stochastic Label Refinery: Toward Better Target Label Distribution." In *2020 25th International Conference on Pattern Recognition (ICPR)*, pp. 9115-9121. IEEE, 2021.

[26] Fujioka, Takuya, Dario Bertero, Takeshi Homma, and Kenji Nagamatsu. "Addressing Ambiguity of Emotion Labels Through Meta-Learning." *arXiv preprint arXiv:1911.02216* (2019).

[27] Gray, Katie LH, Wendy J. Adams, Nicholas Hedger, Kristiana E. Newton, and Matthew Garner. "Faces and awareness: low-level, not emotional factors determine perceptual dominance." *Emotion* 13, no. 3 (2013): 537.

[28] Haber, Nick, Catalin Voss, Azar Fazel, Terry Winograd, and Dennis P. Wall. "A practical approach to real-time neutral feature subtraction for facial expression recognition." In *2016 IEEE Winter Conference on Applications of Computer Vision (WACV)*, pp. 1-9. IEEE, 2016.

[29] Haber, Nick, Catalin Voss, and Dennis Wall. "Making emotions transparent: Google Glass helps autistic kids understand facial expressions through augmented-reaiity therapy." *IEEE Spectrum* 57, no. 4 (2020): 46-52.

[30] Halim, Abbas, Garberson Ford, Stuart Liu-Mayo, Eric Glover, and Dennis P. Wall. "Multi-modular AI Approach to Streamline Autism Diagnosis in Young Children." *Scientific Reports (Nature Publisher Group)* 10, no. 1 (2020).

[31] He, Kaiming, Xiangyu Zhang, Shaoqing Ren, and Jian Sun. "Deep residual learning for image recognition." In *Proceedings of the IEEE conference on computer vision and pattern recognition*, pp. 770-778. 2016.

[32] Hirth, Matthias, Tobias Hoßfeld, and Phuoc Tran-Gia. "Anatomy of a crowdsourcing platform-using the example of microworkers. com." In *2011 Fifth international conference on innovative mobile and internet services in ubiquitous computing*, pp. 322-329. IEEE, 2011.

[33] Hupont, Isabelle, Sandra Baldassarri, Eva Cerezo, and Rafael Del-Hoyo. "Advanced Human Affect Visualization." In *2013 IEEE International Conference on Systems, Man, and Cybernetics*, pp. 2700-2705. IEEE, 2013.

[34] Jerauld, Robert. "Wearable emotion detection and feedback system." U.S. Patent 9,019,174, issued April 28, 2015.

[35] Kalantarian, Haik, Khaled Jedoui, Peter Washington, and Dennis P. Wall. "A mobile game for automatic emotion-labeling of images." *IEEE transactions on games* 12, no. 2 (2018): 213-218.

[36] Kalantarian, Haik, Khaled Jedoui, Peter Washington, Qandeel Tariq, Kaiti Dunlap, Jessey Schwartz, and Dennis P. Wall. "Labeling images with facial emotion and the potential for pediatric healthcare." *Artificial intelligence in medicine* 98 (2019): 77-86.

[37] Kalantarian, Haik, Khaled Jedoui, Kaitlyn Dunlap, Jessey Schwartz, Peter Washington, Arman Husic, Qandeel Tariq, Michael Ning, Aaron Kline, and Dennis Paul Wall. "The performance of emotion classifiers for children with parent-reported autism: quantitative feasibility study." *JMIR mental health* 7, no. 4 (2020): e13174.

[38] Kalantarian, Haik, Peter Washington, Jessey Schwartz, Jena Daniels, Nick Haber, and Dennis P. Wall. "Guess what?." *Journal of healthcare informatics research* 3, no. 1 (2019): 43-66.

[39] Kalantarian, Haik, Peter Washington, Jessey Schwartz, Jena Daniels, Nick Haber, and Dennis Wall. "A gamified mobile system for crowdsourcing video for autism research." In *2018 IEEE international conference on healthcare informatics (ICHI)*, pp. 350-352. IEEE, 2018.

[40] Kairam, Sanjay, and Jeffrey Heer. "Parting crowds: Characterizing divergent interpretations in crowdsourced annotation tasks." In *Proceedings of the 19th ACM Conference on Computer-Supported Cooperative Work & Social Computing*, pp. 1637-1648. 2016.

[41] Kaiser, Robin, and Karina Oertel. "Emotions in HCI: an affective e-learning system." In *Proceedings of the HCSNet workshop on Use of vision in human-computer interaction-Volume 56*, pp. 105-106. Australian Computer Society, Inc., 2006.

[42] Kaur, Ramandeep, and Sandeep Kautish. "Multimodal sentiment analysis: A survey and comparison." *International Journal of Service Science, Management, Engineering, and Technology (IJSSMET)* 10, no. 2 (2019): 38-58.

[43] Kingma, Diederik P., and Jimmy Ba. "Adam: A method for stochastic optimization." *arXiv preprint arXiv:1412.6980* (2014).

[44] Kline, Aaron, Catalin Voss, Peter Washington, Nick Haber, Hessey Schwartz, Qandeel Tariq, Terry Winograd, Carl Feinstein, and Dennis P. Wall. "Superpower glass." *GetMobile: Mobile Computing and Communications* 23, no. 2 (2019): 35-38.

[45] Korovina, Olga, Marcos Baez, and Fabio Casati. "Reliability of crowdsourcing as a method for collecting emotions labels on pictures." *BMC research notes* 12, no. 1 (2019): 1-6.

[46] Korovina, Olga, Fabio Casati, Radoslaw Nielek, Marcos Baez, and Olga Berestneva. "Investigating crowdsourcing as a method to collect emotion labels for images." In *Extended Abstracts of the 2018 CHI Conference on Human Factors in Computing Systems*, pp. 1-6. 2018.

[47] Kosmicki, J. A., V. Sochat, M. Duda, and D. P. Wall. "Searching for a minimal set of behaviors for autism detection through feature selection-based machine learning." *Translational psychiatry* 5, no. 2 (2015): e514-e514.

[48] Kratzwald, Bernhard, Suzana Ilić, Mathias Kraus, Stefan Feuerriegel, and Helmut Prendinger. "Deep learning for affective computing: Text-based emotion recognition in decision support." *Decision Support Systems* 115 (2018): 24-35.

[49] Lease, Matthew. "On quality control and machine learning in crowdsourcing." *Human Computation* 11, no. 11 (2011).

[50] Leblanc, Emilie, Peter Washington, Maya Varma, Kaitlyn Dunlap, Yordan Penev, Aaron Kline, and Dennis P. Wall. "Feature replacement methods enable reliable home video analysis for machine learning detection of autism." *Scientific reports* 10, no. 1 (2020): 1-11.

[51] Liu, Runpeng, Joseph P. Salisbury, Arshya Vahabzadeh, and Ned T. Sahin. "Feasibility of an autism-focused augmented reality smartglasses system for social communication and behavioral coaching." *Frontiers in pediatrics* 5 (2017): 145.

[52] LoBue, Vanessa, Lewis Baker, and Cat Thrasher. "Through the eyes of a child: Preschoolers' identification of emotional expressions from the child affective facial expression (CAFE) set." *Cognition and Emotion* 32, no. 5 (2018): 1122-1130.

[53] LoBue, Vanessa, and Cat Thrasher. "The Child Affective Facial Expression (CAFE) set: Validity and reliability from untrained adults." *Frontiers in psychology* 5 (2015): 1532.

[54] Lotfian, Reza, and Carlos Busso. "Over-sampling emotional speech data based on subjective evaluations provided by multiple individuals." *IEEE Transactions on Affective Computing* (2019).

[55] Magdin, Martin, and F. Prikler. "Real time facial expression recognition using webcam and SDK affectiva." *IJIMAI* 5, no. 1 (2018): 7-15.

[56] McDuff, Daniel, Rana Kaliouby, Thibaud Senechal, May Amr, Jeffrey Cohn, and Rosalind Picard. "Affectiva-mit facial expression dataset (am-fed): Naturalistic and spontaneous facial expressions collected." In *Proceedings of the IEEE Conference on Computer Vision and Pattern Recognition Workshops*, pp. 881-888. 2013.

[57] McFarland, Dennis J., Muhammad A. Parvaz, William A. Sarnacki, Rita Z. Goldstein, and Jonathan R. Wolpaw. "Prediction of subjective ratings of emotional pictures by EEG features." *Journal of neural engineering* 14, no. 1 (2016): 016009.

[58] Mower, Emily, Maja J. Matarić, and Shrikanth Narayanan. "A framework for automatic human emotion classification using emotion profiles." *IEEE Transactions on Audio, Speech, and Language Processing* 19, no. 5 (2010): 1057-1070.

[59] Mower, Emily, Angeliki Metallinou, Chi-Chun Lee, Abe Kazemzadeh, Carlos Busso, Sungbok Lee, and Shrikanth Narayanan. "Interpreting ambiguous emotional expressions." In *2009 3rd International Conference on Affective Computing and Intelligent Interaction and Workshops*, pp. 1-8. IEEE, 2009.

[60] Nguyen, Binh T., Minh H. Trinh, Tan V. Phan, and Hien D. Nguyen. "An efficient real-time emotion detection using camera and facial landmarks." In *2017 seventh international conference on information science and technology (ICIST)*, pp. 251-255. IEEE, 2017.

[61] Ning, Michael, Jena Daniels, Jessey Schwartz, Kaitlyn Dunlap, Peter Washington, Haik Kalantarian, Michael Du, and Dennis P. Wall. "Identification and quantification of gaps in access to autism resources in the United States: an infodemiological study." *Journal of medical Internet research* 21, no. 7 (2019): e13094.

[62] Nicolaou, Mihalis A., Hatice Gunes, and Maja Pantic. "Continuous prediction of spontaneous affect from multiple cues and modalities in valence-arousal space." *IEEE Transactions on Affective Computing* 2, no. 2 (2011): 92-105.

[63] Paolacci, Gabriele, Jesse Chandler, and Panagiotis G. Ipeirotis. "Running experiments on amazon mechanical turk." *Judgment and Decision making* 5, no. 5 (2010): 411-419.

[64] Parthasarathy, Srinivas, and Carlos Busso. "Jointly Predicting Arousal, Valence and Dominance with Multi-Task Learning." In *Interspeech*, vol. 2017, pp. 1103-1107. 2017.

[65] Peterson, Joshua C., Ruairidh M. Battleday, Thomas L. Griffiths, and Olga Russakovsky. "Human uncertainty makes classification more robust." In *Proceedings of the IEEE/CVF International Conference on Computer Vision*, pp. 9617-9626. 2019.

[66] Pioggia, Giovanni, Roberta Igliozzi, Marcello Ferro, Arti Ahluwalia, Filippo Muratori, and Danilo De Rossi. "An android for enhancing social skills and emotion recognition in people with autism." *IEEE Transactions on Neural Systems and Rehabilitation Engineering* 13, no. 4 (2005): 507-515.

[67] Poria, Soujanya, Erik Cambria, and Alexander Gelbukh. "Deep convolutional neural network textual features and multiple kernel learning for utterance-level multimodal sentiment analysis." In *Proceedings of the 2015 conference on empirical methods in natural language processing*, pp. 2539-2544. 2015.

[68] Poria, Soujanya, Erik Cambria, Newton Howard, Guang-Bin Huang, and Amir Hussain. "Fusing audio, visual and textual clues for sentiment analysis from multimodal content." *Neurocomputing* 174 (2016): 50-59.

[69] Rizos, Georgios, and Björn W. Schuller. "Average Jane, Where Art Thou?–Recent Avenues in Efficient Machine Learning Under Subjectivity Uncertainty." In *International Conference on Information Processing and Management of Uncertainty in Knowledge-Based Systems*, pp. 42-55. Springer, Cham, 2020.

[70] Rodrigues, Filipe, and Francisco Pereira. "Deep learning from crowds." In *Proceedings of the AAAI Conference on Artificial Intelligence*, vol. 32, no. 1. 2018.

[71] Sharma, Mukta, Anand Singh Jalal, and Aamir Khan. "Emotion recognition using facial expression by fusing key points descriptor and texture features." *Multimedia Tools and Applications* 78, no. 12 (2019): 16195-16219.

[72] Smitha, Kavallur Gopi, and A. Prasad Vinod. "Facial emotion recognition system for autistic children: a feasible study based on FPGA implementation." *Medical & biological engineering & computing* 53, no. 11 (2015): 1221-1229.

[73] Stappen, Lukas, Alice Baird, Erik Cambria, and Björn W. Schuller. "Sentiment analysis and topic recognition in video transcriptions." *IEEE Intelligent Systems* 36, no. 2 (2021): 88-95.

[74] Tahir, Madiha, Abdallah Tubaishat, Feras Al-Obeidat, Babar Shah, Zahid Halim, and Muhammad Waqas. "A novel binary chaotic genetic algorithm for feature selection and its utility in affective computing and healthcare." *Neural Computing and Applications* (2020): 1-22.

[75] Tao, Jianhua, and Tieniu Tan. "Affective computing: A review." In *International Conference on Affective computing and intelligent interaction*, pp. 981-995. Springer, Berlin, Heidelberg, 2005.

[76] Tariq, Qandeel, Scott Lanyon Fleming, Jessey Nicole Schwartz, Kaitlyn Dunlap, Conor Corbin, Peter Washington, Haik Kalantarian, Naila Z. Khan, Gary L. Darmstadt, and Dennis Paul Wall. "Detecting developmental delay and autism through machine learning models using home videos of Bangladeshi children: Development and validation study." *Journal of medical Internet research* 21, no. 4 (2019): e13822.

[77] Tariq, Qandeel, Jena Daniels, Jessey Nicole Schwartz, Peter Washington, Haik Kalantarian, and Dennis Paul Wall. "Mobile detection of autism through machine learning on home video: A development and prospective validation study." *PLoS medicine* 15, no. 11 (2018): e1002705.

[78] Thiam, Patrick, Sascha Meudt, Markus Kächele, Günther Palm, and Friedhelm Schwenker. "Detection of emotional events utilizing support vector methods in an active learning HCI scenario." In *Proceedings of the 2014 workshop on emotion representation and modelling in human-computer-interaction-systems*, pp. 31-36. 2014.

[79] Thiel, Christian. "Classification on soft labels is robust against label noise." In *International Conference on Knowledge-Based and Intelligent Information and Engineering Systems*, pp. 65-73. Springer, Berlin, Heidelberg, 2008.

[80] Turing, Alan M. "Computing machinery and intelligence." In *Parsing the turing test*, pp. 23-65. Springer, Dordrecht, 2009.

[81] Uma, Alexandra, Tommaso Fornaciari, Dirk Hovy, Silviu Paun, Barbara Plank, and Massimo Poesio. "A Case for Soft Loss Functions." In *Proceedings of the AAAI Conference on Human Computation and Crowdsourcing*, vol. 8, no. 1, pp. 173-177. 2020.

[82] Villon, Olivier, and Christine Lisetti. "Toward recognizing individual's subjective emotion from physiological signals in practical application." In *Twentieth IEEE International Symposium on Computer-Based Medical Systems (CBMS'07)*, pp. 357-362. IEEE, 2007.

[83] Völkel, Sarah Theres, Julia Graefe, Ramona Schödel, Renate Häuslschmid, Clemens Stachl, Quay Au, and Heinrich Hussmann. "I Drive My Car and My States Drive Me: Visualizing Driver's Emotional and Physical States." In *Adjunct Proceedings of the 10th International Conference on Automotive User Interfaces and Interactive Vehicular Applications*, pp. 198-203. 2018.

[84] Voss, Catalin, Jessey Schwartz, Jena Daniels, Aaron Kline, Nick Haber, Peter Washington, Qandeel Tariq et al. "Effect of wearable digital intervention for improving socialization in children with autism spectrum disorder: a randomized clinical trial." *JAMA pediatrics* 173, no. 5 (2019): 446-454.

[85] Voss, Catalin, Peter Washington, Nick Haber, Aaron Kline, Jena Daniels, Azar Fazel, Titas De et al. "Superpower glass: delivering unobtrusive real-time social cues in wearable systems." In *Proceedings of the 2016 ACM International Joint Conference on Pervasive and Ubiquitous Computing: Adjunct*, pp. 1218-1226. 2016.

[86] Wall, Dennis Paul, J. Kosmicki, T. F. Deluca, E. Harstad, and Vincent Alfred Fusaro. "Use of machine learning to shorten observation-based screening and diagnosis of autism." *Translational psychiatry* 2, no. 4 (2012): e100-e100.

[87] Washington, Peter, Emilie Leblanc, Kaitlyn Dunlap, Yordan Penev, Aaron Kline, Kelley Paskov, Min Woo Sun et al. "Precision Telemedicine through Crowdsourced Machine Learning: Testing Variability of Crowd Workers for Video-Based Autism Feature Recognition." *Journal of personalized medicine* 10, no. 3 (2020): 86.

[88] Washington, Peter, Natalie Park, Parishkrita Srivastava, Catalin Voss, Aaron Kline, Maya Varma, Qandeel Tariq et al. "Data-driven diagnostics and the potential of mobile artificial intelligence for digital therapeutic phenotyping in computational psychiatry." *Biological Psychiatry: Cognitive Neuroscience and Neuroimaging* (2019).

[89] Washington, Peter, Haik Kalantarian, Jack Kent, Arman Husic, Aaron Kline, Emilie Leblanc, Cathy Hou et al. "Training an Emotion Detection Classifier using Frames from a Mobile Therapeutic Game for Children with Developmental Disorders." *arXiv preprint arXiv:2012.08678* (2020).

[90] Washington, Peter, Haik Kalantarian, Qandeel Tariq, Jessey Schwartz, Kaitlyn Dunlap, Brianna Chrisman, Maya Varma et al. "Validity of online screening for autism: crowdsourcing study comparing paid and unpaid diagnostic tasks." *Journal of medical Internet research* 21, no. 5 (2019): e13668.

[91] Washington, Peter, Aaron Kline, Onur Cezmi Mutlu, Emilie Leblanc, Cathy Hou, Nate Stockham, Kelley Paskov, Brianna Chrisman, and Dennis P. Wall. "Activity Recognition with Moving Cameras and Few Training Examples: Applications for Detection of Autism-Related Headbanging." *arXiv preprint arXiv:2101.03478* (2021).

[92] Washington, Peter, Emilie Leblanc, Kaitlyn Dunlap, Yordan Penev, Maya Varma, Jae-Yoon Jung, Brianna Chrisman et al. "Selection of trustworthy crowd workers for telemedical diagnosis of pediatric autism spectrum disorder." In *BIOCOMPUTING 2021: Proceedings of the Pacific Symposium*, pp. 14-25. 2020.

[93] Washington, Peter, Kelley Marie Paskov, Haik Kalantarian, Nathaniel Stockham, Catalin Voss, Aaron Kline, Ritik Patnaik et al. "Feature selection and dimension reduction of social autism data." In *Pac Symp Biocomput*, vol. 25, pp. 707-718. 2020.

[94] Washington, Peter, Qandeel Tariq, Emilie Leblanc, Brianna Chrisman, Kaitlyn Dunlap, Aaron Kline, Haik Kalantarian et al. "Crowdsourced privacy-preserved feature tagging of short home videos for machine learning ASD detection." *Scientific reports* 11, no. 1 (2021): 1-11.

[95] Washington, Peter, Catalin Voss, Nick Haber, Serena Tanaka, Jena Daniels, Carl Feinstein, Terry Winograd, and Dennis Wall. "A wearable social interaction aid for children with autism." In *Proceedings of the 2016 CHI Conference Extended Abstracts on Human Factors in Computing Systems*, pp. 2348-2354. 2016.

[96] Washington, Peter, Catalin Voss, Aaron Kline, Nick Haber, Jena Daniels, Azar Fazel, Titas De, Carl Feinstein, Terry Winograd, and Dennis Wall. "SuperpowerGlass: a wearable aid for the at-home therapy of children with autism." *Proceedings of the ACM on interactive, mobile, wearable and ubiquitous technologies* 1, no. 3 (2017): 1-22.

[97] Washington, Peter, Serena Yeung, Bethany Percha, Nicholas Tatonetti, Jan Liphardt, and Dennis P. Wall. "Achieving trustworthy biomedical data solutions." In *BIOCOMPUTING 2021: Proceedings of the Pacific Symposium*, pp. 1-13. 2020.

[98] White, Susan W., Lynn Abbott, Andrea Trubanova Wieckowski, Nicole N. Capriola-Hall, Sherin Aly, and Amira Youssef. "Feasibility of automated training for facial emotion expression and recognition in autism." *Behavior therapy* 49, no. 6 (2018): 881-888.

[99] Yang, Zhao, Tie Liu, Jiehao Liu, Li Wang, and Sai Zhao. "A Novel Soft Margin Loss Function for Deep Discriminative Embedding Learning." *IEEE Access* 8 (2020): 202785-202794.

[100] Yannakakis, Georgios N. "Enhancing health care via affective computing." (2018).

[101] Yin, Da, Xiao Liu, Xiuyu Wu, and Baobao Chang. "A soft label strategy for target-level sentiment classification." In *Proceedings of the Tenth Workshop on Computational Approaches to Subjectivity, Sentiment and Social Media Analysis*, pp. 6-15. 2019.

[102] Yu, Liang-Chih, Jin Wang, K. Robert Lai, and Xue-jie Zhang. "Predicting valence-arousal ratings of words using a weighted graph method." In *Proceedings of the 53rd Annual Meeting of the Association for Computational Linguistics and the 7th International Joint Conference on Natural Language Processing (Volume 2: Short Papers)*, pp. 788-793. 2015.

[103] Zeng, Zhihong, Jilin Tu, Ming Liu, Thomas S. Huang, Brian Pianfetti, Dan Roth, and Stephen Levinson. "Audio-visual affect recognition." *IEEE Transactions on multimedia* 9, no. 2 (2007): 424-428.

[104] Zhang, Lin, Steffen Walter, Xueyao Ma, Philipp Werner, Ayoub Al-Hamadi, Harald C. Traue, and Sascha Gruss. ""BioVid Emo DB": A multimodal database for emotion analyses validated by subjective ratings." In *2016 IEEE Symposium Series on Computational Intelligence (SSCI)*, pp. 1-6. IEEE, 2016.

[105] Zhang, Xiao, Wenzhong Li, Xu Chen, and Sanglu Lu. "Moodexplorer: Towards compound emotion detection via smartphone sensing." *Proceedings of the ACM on Interactive, Mobile, Wearable and Ubiquitous Technologies* 1, no. 4 (2018): 1-30.

[106] Zhao, Sicheng, Hongxun Yao, and Xiaolei Jiang. "Predicting continuous probability distribution of image emotions in valence-arousal space." In *Proceedings of the 23rd ACM international conference on Multimedia*, pp. 879-882. 2015.

[107] Zhou, Ying, Xuefeng Liang, Yu Gu, Yifei Yin, and Longshan Yao. "Multi-Classifier Interactive Learning for Ambiguous Speech Emotion Recognition." *arXiv preprint arXiv:2012.05429* (2020).